\documentclass[letterpaper, 10 pt, conference]{ieeeconf}  

\IEEEoverridecommandlockouts                              

\usepackage{amsmath,amssymb,amsfonts}
\usepackage{algorithmic}
\usepackage{graphicx}
\usepackage{textcomp}
\usepackage[table]{xcolor}
\usepackage{caption}
\usepackage{multirow}
\usepackage{footnote}
\usepackage{hyperref}
\usepackage{array}
\usepackage{amssymb}
\usepackage{todonotes}
\usepackage[pscoord]{eso-pic}
\usepackage{multirow}
\usepackage{mathtools}
\usepackage{cite}
\DeclareMathOperator*{\argmax}{arg\,max}

\title{\LARGE \bf
Vision based Crop Row Navigation under Varying Field Conditions in Arable Fields
}

\author{Rajitha de Silva$^{1}$, Grzegorz Cielniak$^{2}$ and Junfeng Gao$^{3}$
\thanks{This work was supported by Lincoln Agri-Robotics as part of the Expanding Excellence in England (E3) Programme.}
\thanks{$^{1}$Rajitha de Silva, $^{2}$Grzegorz Cielniak and $^{3}$Junfeng Gao are with Lincoln Agri-Robotics Centre, Lincoln Institute for Agri-Food Technology, University of Lincoln, UK
        {\tt\small $^{1}$rajitha@ieee.org, $^{2}$gcielniak@lincoln.ac.uk, $^{3}$jugao@lincoln.ac.uk}}%
}

\begin{document}

\maketitle

\begin{abstract}
Accurate crop row detection is often challenged by the varying field conditions present in real-world arable fields. Traditional colour based segmentation is unable to cater for all such variations. The lack of comprehensive datasets in agricultural environments limits the researchers from developing robust segmentation models to detect crop rows. We present a dataset for crop row detection with 11 field variations from Sugar Beet and Maize crops. We also present a novel crop row detection algorithm for visual servoing in crop row fields. Our algorithm can detect crop rows against varying field conditions such as curved crop rows, weed presence, discontinuities, growth stages, tramlines, shadows and light levels. Our method only uses RGB images from a front-mounted camera on a Husky robot to predict crop rows. Our method outperformed the classic colour based crop row detection baseline. Dense weed presence within inter-row space and discontinuities in crop rows were the most challenging field conditions for our crop row detection algorithm. Our method can detect the end of the crop row and navigate the robot towards the headland area when it reaches the end of the crop row.

\end{abstract}

\section{Introduction}
The sustainability of agri-food production is an important challenge when catering for the growing demand in the global food supply. While the usage of tractors and human labour has been the usual agricultural practice for many years, the usage of robots in agriculture has become an essential need for the future of sustainable agriculture. The state-of-the-art technology could safely guide robots in arable fields without damaging the crops. However, sensors needed for such robots such as Real Time Kinematic Global Positioning System (RTK-GPS) and LiDAR are expensive. The development of cheaper technological alternatives would enable faster adoption of these technologies.

Vision sensors are an alternative to expensive sensors used in navigation. However, the benefit of having vision sensors could only be fully realized with accurate computer vision algorithms for vision based navigation \cite{oliveira2021advances}. Crop row detection is an important step of such vision based navigation systems to accurately guide the robot to follow a crop row. Earlier work on crop row detection used colour based image segmentation to generate crop row masks followed by some image processing techniques to obtain the line parameters of the detected crop row \cite{bonadies2019overview}. A new generation of crop row detection algorithms was revealed with the advancements in deep learning for image segmentation \cite{adhikari2020deep, pang2020improved, bah2019crownet}. Most of these crop row detection algorithms are only tested under limited field variations. Some of the algorithms are robust against weed presence, varying growth stages, discontinuities and shadows \cite{ji2011crop, fue2020evaluation}. However, most of these algorithms are only focused on one field variation rather than many other field conditions that may arise in an arable field.

The objective of our work was to develop a robust crop row detection algorithm for visual servoing using deep learning based segmentation for accurate detection. The main contributions of our work are summarized as follows:

\begin{itemize}
  \item An extended dataset for crop row detection with multiple crops.
  \item A novel crop row scanning algorithm which accurately localise the central crop row from the segmentation mask of the crop row.
  \item Evaluation of crop row detection performance under varying field conditions in Sugar Beet and Maize fields.
  \item A visual servoing based crop row navigation algorithm which could follow through and exit the crop row at its endpoint.

\end{itemize}

\section{Related Work}
\label{sec:bgr}
Crop row detection for vision based navigation is perceived as a 2-stage process \cite{bonadies2019overview}. The first stage involves image processing to segment images belonging to the crop. The second stage determines the line parameters for the crop rows from the segmented crop row mask. Crop segmentation is often realized colour based segmentation approaches \cite{ahmadi2020visual, romeo2012crop, guerrero2013automatic}. Excess Green Index (ExG) \cite{woebbecke1995color}, living tissue indicator and vegetation index \cite{bakker2008vision, montalvo2012automatic} are some of the popular colour based segmentation methods. The advancements in deep learning could be leveraged to obtain a cleaner crop row mask that needs less post-processing before predicting the line parameters. Such methods can predict consistent crop row masks under varying field conditions \cite{adhikari2020deep, bah2019crownet, emmi2022toward, fawakherji2019crop}. We have verified the capability of such deep learning based segmentation methods in detecting crop rows under varying field conditions in our previous work \cite{de2022towards}. The initial segmentation provides a binary image representation of pixels belonging to the crop. That segmentation mask must be further processed before attempting to determine the line parameters for crop rows. The Hough transform was used in predicting line parameters of crop rows \cite{winterhalter2018crop, GAO201843}. A limitation of such Hough transform based crop row detection is that the Hough transform parameters must be tuned for each type of plant and field conditions. Therefore, using the Hough transform limits the algorithm's ability for crop row detection under varying field conditions. 

Winterhalter et al. \cite{winterhalter2021localization} have used GNSS (Global Navigation Satellite System) labelled maps to detect the end of field for switching from one crop row to another. They highlight the advantage of GNSS for end-of-field detection. However, such an advantage comes at the expensive cost of GNSS hardware. A vision based end of row (EOR) detector would be better to reduce the overall cost of agricultural robots. Researchers have used monocular cameras with variable field of view (FOV) to accurately predict crop rows near the EOR \cite{xue2012variable}. However, early EOR detection allows our method to plan a short exit path to exit the crop row without needing a variable FOV camera.

Ahmadi et al. \cite{ahmadi2021towards} developed a vision based navigation system for arable fields that uses colour based segmentation for crop row mask prediction. Their crop row line parameter estimation was based on the least square fitting of detected crop centres. They have tested their system in 5 different crop fields with various canopy types. We choose their method as a baseline for our work to compare the effect of colour based segmentation against deep learning based segmentation.

\section{Dataset}
\label{sec:dt}
The images in the CRDLDv2.1\footnote{The dataset can be accessed with the following link: \textbf{\url{https://github.com/JunfengGaolab/CropRowDetection}}.} (Crop Row Detection Lincoln Dataset) dataset presented in this paper was captured from an Intel Realsense D435i camera mounted on the front of a Husky robot. CRDLDv2.1 is an update on CRDLDv2.0 \cite{desilva22deep} with addition of Maize crop data into the dataset. The dataset contains images from a Sugar Beet field and early growth stage images from a Maize field.

\subsection{Data Categories}
\begin{figure}[t]
\centering
\captionsetup{justification=centering}
\includegraphics[scale=0.28]{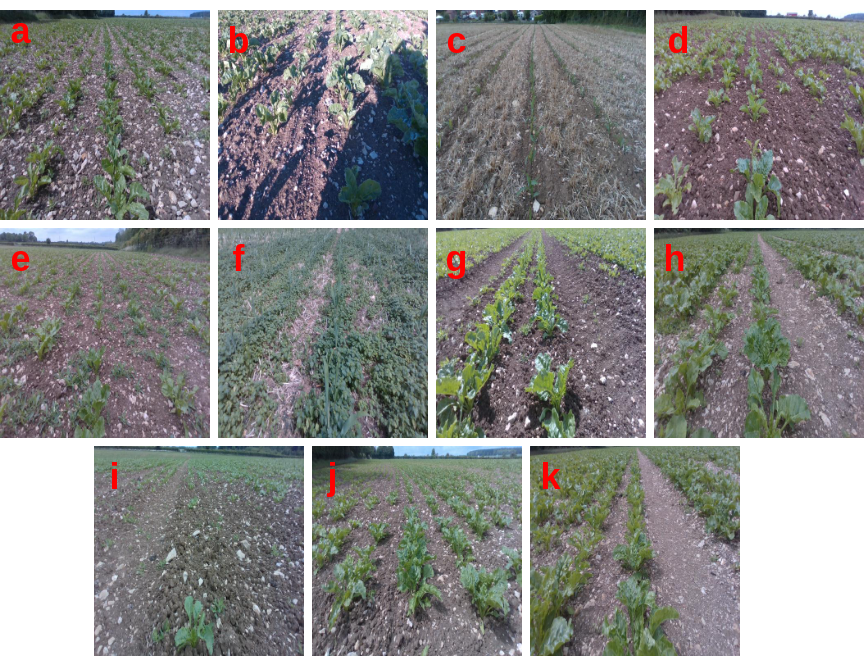}
\caption{Samples from 11 Field Variations}
\label{fig:cat}
\end{figure}

We have identified 11 field variations present in the captured data as shown in Figure \ref{fig:cat}. The identified field variations are explained in Table \ref{tab:cat}. The captured Sugar Beet data were classified into 43 data classes representing various possible combinations of the 11 identified field variations. The captured Maize data were classified into 7 data classes representing possible combinations from 6 of the identified field variations. The dataset is comprised of a total of 50 data classes across 2 crops with 11 field variations. The 50 variations are listed in Table \ref{tab:cls} where classes 1-43 are from the Sugar Beet crop and classes 44-50 are from the Maize crop. The inter-row space of crop rows in Sugar Beet field often contain rocks as visible in examples $a$, $e$ and $i$ of Figure \ref{fig:cat}. The inter-row space of the Maize field is populated with dead plant residue after harvesting from the previous crop season as seen in example $c$ of Figure \ref{fig:cat}. This difference in crop background is also an interesting variation although it is not considered a field variation.

\begin{table}[t]
\centering
\caption{Data Categories}
\begin{center}
\begin{tabular}{|p{0.04\linewidth} | p{0.21\linewidth} | p{0.55\linewidth}|}
\hline
\textbf{ID} & \textbf{Data Category} & \textbf{Description}\\
\hline
a & Horizontal Shadow & Shadow falls perpendicular to the direction of the crop row \\ 
\hline
b & Front Shadow & Shadow of the robot falling on the image captured by the camera \\ 
\hline
c & Small Crops & Crop rows at early growth stages (Up to 4 unfolded leaves) \\ 
\hline
d & Large Crops & Presence of one or many largely grown crops (more than 4 unfolded leaves) within the crop row \\ 
\hline
e & Sparse Weed & Sparsely grown weed scattered between the crop rows \\ 
\hline
f & Dense Weed & Weed grown densely among the crop rows where the inter-row space is completely covered\\ 
\hline
g & Sunny & Crop row data captured in sunny weather \\ 
\hline
h & Cloudy & Crop row data captured in cloudy weather \\ 
\hline
i & Discontinuities & Missing plants in the crop row which leads to discontinuities in crop row \\ 
\hline
j & Slope/ Curve & Images captured while the crop row is not in a flat farmland or where crop rows are not straight lines \\ 
\hline
k & Tyre Tracks & Tyre tracks from tramlines running through the field \\ 
\hline
\end{tabular}
\label{tab:cat}
\end{center}
\end{table}

\begin{table*}[t]
\centering
\caption{Data Classes (The Numbers given in this table correspond to the indexes of combined field variations from different data categories. Sugar Beet:1-43 Maize:44-50)}
\begin{center}
\begin{tabular}{|p{0.15\linewidth}| p{0.04\linewidth} | p{0.04\linewidth} | p{0.04\linewidth} | p{0.04\linewidth} | p{0.04\linewidth} | p{0.04\linewidth} | p{0.04\linewidth} | p{0.04\linewidth} | p{0.04\linewidth} | p{0.04\linewidth} |}
\hline
\textbf{Data Category} & \multicolumn{1}{|c|}a & \multicolumn{1}{|c|}b & \multicolumn{1}{|c|}c & \multicolumn{1}{|c|}d & \multicolumn{1}{|c|}e & \multicolumn{1}{|c|}f & \multicolumn{1}{|c|}g & \multicolumn{1}{|c|}h & \multicolumn{1}{|c|}i & \multicolumn{1}{|c|}j\\
\hline
Horizontal Shadow (a) & \multicolumn{1}{|c|}{1} & \cellcolor{black!25} & \cellcolor{black!25} & \cellcolor{black!25} & \cellcolor{black!25} & \cellcolor{black!25} & \cellcolor{black!25} & \textbf{\cellcolor{black!25}} & \cellcolor{black!25} & \cellcolor{black!25}  \\ 
\hline
Front Shadow (b) & \cellcolor{black!25} & \multicolumn{1}{|c|}7 & \cellcolor{black!25} & \cellcolor{black!25} & \cellcolor{black!25} & \cellcolor{black!25} & \cellcolor{black!25} & \cellcolor{black!25} & \cellcolor{black!25} & \cellcolor{black!25} \\ 
\hline
Small Crops (c) & \multicolumn{1}{|c|}2 & \multicolumn{1}{|c|}8 & 11, 44 & \cellcolor{black!25} & \multicolumn{1}{|c|}{47} & \cellcolor{black!25} & \cellcolor{black!25} & \cellcolor{black!25} & \multicolumn{1}{|c|}{45} & \multicolumn{1}{|c|}{46} \\ 
\hline
Large Crops (d) & \multicolumn{1}{|c|}3 & \multicolumn{1}{|c|}9 & \multicolumn{1}{|c|}{12} & 20, 48 & \multicolumn{2}{|c|}{50} & \cellcolor{black!25} & \cellcolor{black!25} & \multicolumn{1}{|c|}{49} & \cellcolor{black!25} \\ 
\hline
Sparse Weed (e) & \cellcolor{black!25} & \cellcolor{black!25} & \multicolumn{1}{|c|}{13} & \multicolumn{1}{|c|}{21} & \cellcolor{black!25} & \cellcolor{black!25} & \cellcolor{black!25} & \cellcolor{black!25} & \cellcolor{black!25} & \cellcolor{black!25} \\ 
\hline
Dense Weed (f) & \cellcolor{black!25} & \cellcolor{black!25} & \multicolumn{1}{|c|}{14} & \multicolumn{1}{|c|}{22} & \cellcolor{black!25} & \cellcolor{black!25} & \cellcolor{black!25} & \cellcolor{black!25} & \cellcolor{black!25} & \cellcolor{black!25} \\ 
\hline
Sunny (g) & \cellcolor{black!25} & \cellcolor{black!25} & \multicolumn{1}{|c|}{15} & \multicolumn{1}{|c|}{23} & \multicolumn{1}{|c|}{28} & \multicolumn{1}{|c|}{32} & \cellcolor{black!25} & \cellcolor{black!25} & \textbf{\cellcolor{black!25}} & \cellcolor{black!25} \\ 
\hline
Cloudy (h) & \cellcolor{black!25} & \cellcolor{black!25} & \multicolumn{1}{|c|}{16} & \multicolumn{1}{|c|}{24} & \multicolumn{1}{|c|}{29} & \multicolumn{1}{|c|}{33} & \cellcolor{black!25} & \cellcolor{black!25} & \cellcolor{black!25} & \cellcolor{black!25} \\ 
\hline
Discontinuities (i) & \multicolumn{1}{|c|}4 & \multicolumn{1}{|c|}{10} & \multicolumn{1}{|c|}{17} & \multicolumn{1}{|c|}{25} & \multicolumn{1}{|c|}{30} & \multicolumn{1}{|c|}{34} & \multicolumn{1}{|c|}{36} & \multicolumn{1}{|c|}{39} & \cellcolor{black!25} & \cellcolor{black!25} \\ 
\hline
Slope/ Curve (j) & \multicolumn{1}{|c|}5 & \cellcolor{black!25} & \multicolumn{1}{|c|}{18} & \multicolumn{1}{|c|}{26} & \multicolumn{1}{|c|}{31} & \multicolumn{1}{|c|}{35} & \multicolumn{1}{|c|}{37} & \multicolumn{1}{|c|}{40} & \multicolumn{1}{|c|}{42} & \cellcolor{black!25} \\ 
\hline
Tyre Tracks (k) & \multicolumn{1}{|c|}6 & \cellcolor{black!25} & \multicolumn{1}{|c|}{19} & \multicolumn{1}{|c|}{27} & \cellcolor{black!25} & \cellcolor{black!25} & \multicolumn{1}{|c|}{38} & \multicolumn{1}{|c|}{41} & \multicolumn{1}{|c|}{43} & \cellcolor{black!25} \\ 
\hline
\end{tabular}
\label{tab:cls}
\end{center}
\end{table*}

\subsection{Data Preprocessing and Annotation}
\begin{figure}
\centering
\captionsetup{justification=centering}
\includegraphics[scale=0.22]{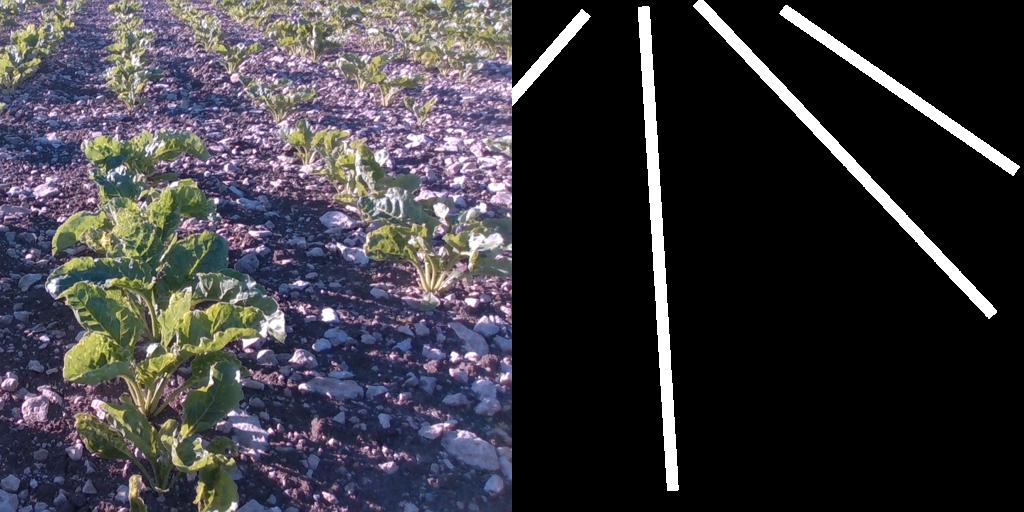}
\caption{Sample image and respective ground truth label mask}
\label{fig:lbl}
\end{figure}

The RGB images captured by the Realsense camera were resized to 512x512 resolution to be used as the input for the U-Net\cite{ronneberger2015u} Convolutional Neural Network (CNN). The starting point and end point of each crop row were labelled on the MATLAB VideoLabeller tool during the data annotation phase. The curved lines were approximated to a set of straight-line segments. These annotations were further processed to generate ground truth crop row masks where each crop row was represented by a white line with a width of 6 pixels. Figure \ref{fig:lbl} is an example of an RGB and ground truth image pair. The white pixels in the ground truth masks do not represent the spatial occupancy of the crop within each image. The white pixels lines are a representation of the high-level semantic feature known as the "Crop Row". This labelling approach will ensure that the predicted crop row masks will be narrow lines representing the spatial positioning of the crop row rather than predicting the pixels belonging to the crop in each RGB image.

\section{Methodology}
\label{sec:mtd}

\begin{figure}[t]
\centering
\captionsetup{justification=centering}
\includegraphics[scale=0.35]{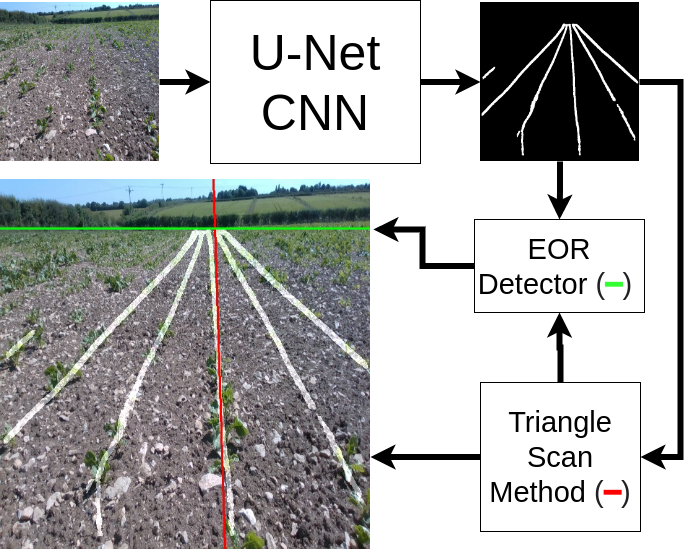}
\caption{Proposed Crop Row Detection Pipeline}
\label{fig:ovl}
\end{figure}

Figure \ref{fig:ovl} illustrates the main steps of the crop row detection architecture of the autonomous navigation pipeline proposed in this paper. An RGB image from the front-mounted camera of the Husky robot is fed into the U-Net CNN. The U-Net predicts the crop row mask of the image which isolates the crop row objects in the image. The detected crop row mask is processed using the triangle scan method to detect the endpoints of the central crop row which the robot must follow. Accurate detection of the central crop row is a vital step in detecting crop rows for visual servoing of a robot. Most of the existing methods detect all the existing rows within an image and then determine the central crop row heuristically. Our method differentiates from those approaches by detecting the central crop row directly from the crop row mask. The subsections \ref{sec:unet} and \ref{sec:tsm} explains the segmentation and crop row detection steps of the proposed crop row detection method. EOR detection is also an important step in developing a vision based navigation system for arable fields. The prediction from the U-Net CNN could also be used to detect the EOR position within an image. The triangle scan method detects a possible instance of EOR when an image closer to the EOR is received. The triangle scan method then triggers the EOR detector which scans the U-Net prediction for EOR position. This process is elaborated in Section \ref{sec:eor}.

\subsection{U-Net CNN}
\label{sec:unet}
The U-Net was trained on the dataset described in Section \ref{sec:dt}. The CNN model was trained with Adam optimizer with Binary Cross Entropy loss function. Model loss function saturated around 40 epochs during the training. Intersection over Union (IoU) is commonly used to measure the success of a CNN in semantic segmentation. However, we have identified that lower IoU in prediction doesn't necessarily reflect lesser capability in crop row detection. The ground truth images used in training the U-Net have crop rows with a width of 6 pixels. The predictions have crop rows of varying widths ranging from 3 pixels to 8 pixels at different points in the crop row. Such variations lead to a lower IoU value. However, the crop row detection performance of our method is unaffected by this as long as the predicted crop row mask consists of an accurate row structure rather than the accurate row width.

\subsection{Triangle Scan Method}
\label{sec:tsm}

The triangle scan method is a 2-step process which determines the central crop row from the U-Net prediction. The first step (Anchor Scan) and the second step (Line Scan) scan for the topmost (Anchor Point) and the lowermost ($P_{r}$) points of the central crop row respectively. The anchor point and point $P_{r}$ always lie on the top and bottom edges of the image respectively. These two steps are explained in subsections \ref{ssec:as} and \ref{ssec:ls}.

\begin{figure}[t]
\centering
\captionsetup{justification=centering}
\includegraphics[scale=0.38]{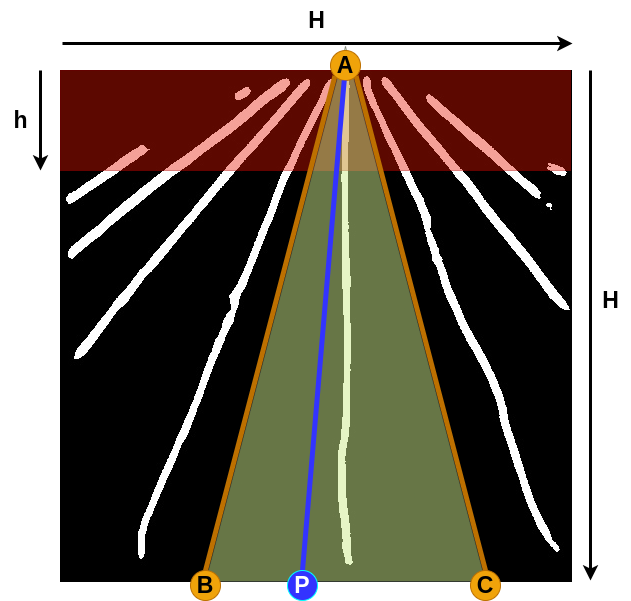}
\caption{Regions of interest for anchor scan and line scan. Anchor Scan ROI: RED, Line Scan ROI: Green}
\label{fig:roi}
\end{figure}

\subsubsection{Anchor Scans}
\label{ssec:as}
Anchor scans stage scans a rectangular region of interest (ROI) of the predicted crop row mask as indicated in red in Figure \ref{fig:roi}. The width of the rectangular ROI is the same as the width of the image while the height of the rectangular ROI is defined by $h$ such that $h=sH$. The height of the input image is $H$ and $s$ is a scaling factor between $0$ and $1$. For our experiments $s$ was set to $0.2$. Equation \ref{eq:1} describes the scanning for the anchor point (A). $I$ is the U-Net prediction and $X$ represents horizontal coordinates within a predefined range in the rectangular ROI. The predefined range (0.2H to 0.7H) for $X$ was experimentally determined by observing the usual anchor point occurrence in the dataset. The anchor point is only validated if the scanner parameter: $\left( \sum_{y=0}^{h} I(X,y) \right)$ meets an experimentally predetermined threshold value. The ROI is shifted down by a distance of $h$ (maximum up to 2 times) if the detected anchor point is invalid. This shifting down helps identify the $A$ when the robot reaches the end of a crop row.

\begin{equation} \label{eq:1}
  A = \argmax \left( \sum_{y=0}^{h} I(X,y) \right) 
\end{equation}

\subsubsection{Line Scans}
\label{ssec:ls}
A triangular ROI is defined in the line scans stage with points A, B (Begin Point) and C (Cease Point) as shown in Figure \ref{fig:roi}. Equation \ref{eq:2} describe the scanning for $P_{r}$ by searching for an instance of a variable point $P$ on $BC$ line which yields the highest pixel sum for $AP$ line.

\begin{equation} \label{eq:2}
  P_{r} = \argmax \Biggl[ \sum_{I_{xy}=A}^{P} I(x,y) \Biggr]_{P=B}^{P=C}
\end{equation}

The maximum recorded offset for $P_{r}$ from A was under 0.2H pixels. The minimum and maximum recorded $P_{r}$ was always above 0.4H pixels and below 0.9H pixels. Therefore, the points $B$ and $C$ are defined by the piecewise function given in \ref{eq:pw}. Points $B$ and $C$ will always lie on the lowermost edge of the image.

\begin{equation} \label{eq:pw}
[B,\: C] = 
    \begin{cases}
        B=0.4H & \text{if } A \leq 0.6H\\
        B=A-0.2H, & \text{if }  A > 0.6H\\
        C=0.9H, & \text{if } A \geq 0.7H\\
        C=A+0.2H, & \text{if }  A<0.7H
    \end{cases}
\end{equation}

\subsection{End of Row Detector}
\label{sec:eor}
The EOR detector is only triggered when the robot reaches the end of a crop row. This could be identified when the rectangular ROI of anchor scans is shifted down. Equation \ref{eq:3} describes the scanning for $EOR$ where $Y$ represents all vertical coordinates in the rectangular ROI and $H$ is the width of the rectangular ROI. The EOR detector scans in rectangular ROI within $[(n-1)h,\: nh]$ range where $n$ is the number of times which the anchor scans ROI shifted $(n_{max} = 2)$.

\begin{equation} \label{eq:3}
  EOR = \argmax \left( \sum_{x=0}^{H} I(x,Y) \right) 
\end{equation}

The successive EOR values after the first detected EOR are updated with a complementary filter to reduce noise in EOR detection. An exit manoeuvre will be executed by the robot when the EOR is detected within the rectangular ROI of $[2h,\: 3h]$ range. The exit manoeuvre is a timed command sequence to guide the robot along a short trajectory to completely exit the crop row. 

\subsection{Visual Servoing Controller}
The forward linear velocity of the robot is always kept constant during the crop row following. The angle that the crop row makes with the vertical direction is $\Delta\theta$ and the positional error of the line from the desired position is $\Delta P_{r}$. A proportional controller was implemented to steer the robot to follow the central crop row. $\Delta\theta$ and $\Delta P_{r}$ are the inputs for the visual servoing controller. The angular velocity output ($\dot{\omega}$) of the controller is described by Equation \ref{eq:pc} where $\alpha$ is the proportional gain and $[w_{1},w_{2}]$ are the contributing weights for angle and displacement errors of detected crop rows.

\begin{equation} \label{eq:pc}
  \dot{\omega} = \alpha \left( w_{1}\Delta\theta + w_{2} \Delta P_{r} \right)
\end{equation}

\subsection{Exit Manoeuvre}
\label{ssec:em}
The angular velocity for robot steering will be generated by $\hat{\omega}(t)$ during the exit manoeuvre. The exit manoeuvre is described by an exponential decay function given in Equation \ref{eq:em}. $\dot{\omega}_{EOR}$ is the $\dot{\omega}$ value when the EOR is first detected within the rectangular ROI of $[2h,\: 3h]$ range and $\lambda$ is a constant that corresponds to the remaining distance of the crop row which the robot has to traverse to exit the crop row. The $\lambda$ may also depend on the type of terrain and linear speed of the robot.

\begin{equation} \label{eq:em}
  \hat{\omega}(t) = \dot{\omega}_{EOR}\times e^{-\lambda t}
\end{equation}

The robot will be brought to a complete halt after $T_e$ seconds from the start of the exit manoeuvre. $T_e$ value could be experimentally determined after observing the average time taken by the $\hat{\omega}(t)$ function to drive the robot out of the crop row. $T_e$ was set to 20 seconds for our simulation.

\section{Results and Discussion}
\label{sec:rslt}

\subsection{Baseline}
Ahmedi et al. \cite{ahmadi2021towards} has developed a vision based navigation scheme that will detect the crop rows and drive the robot using the IBVS(image absed visual servoing) controller \cite{espiau1992new}. They identify the central crop row using least square fitting of crop centers detected using a colour based segmentation approach. Their method shares a similar architecture to ours while using  a colour based segmentation approach in contrast to deep learning approach used by us. They have highlighted the shortcomings of previous colour based crop row detection algorithms placing them at the state-of-the-art in colour based crop row detection. The baseline algorithm was tested using the codes provided by the authors of \cite{ahmadi2021towards} on our dataset.

\subsection{Crop Row Detection Performance Metric}
\label{sec:eps}
The crop row detection performance is evaluated by ($\epsilon$) score described in Equation \ref{eq:ep}. The $\epsilon$ score equally weighs the line detection performance in terms of $\Delta \theta$ and $\Delta P_{r}$. $N$ is the number of images being tested. $\Delta\theta_{max}$ and $\Delta P_{r,max}$ are the maximum detected errors for each $\Delta\theta$ and $\Delta P_{r}$ respectively. The angles are represented in degrees and displacements are expressed in pixels.

\begin{equation} \label{eq:ep}
  \epsilon = 1 - \sum_{i=0}^{N} \frac{1}{2N} \left( \frac{\Delta\theta_{i}}{\Delta\theta_{max}} + \frac{\Delta P_{r,i}}{\Delta P_{r,max}} \right)
\end{equation}

\subsection{Crop Row Detection Evaluation}
\begin{figure}[t]
\centering
\captionsetup{justification=centering}
\includegraphics[scale=0.17]{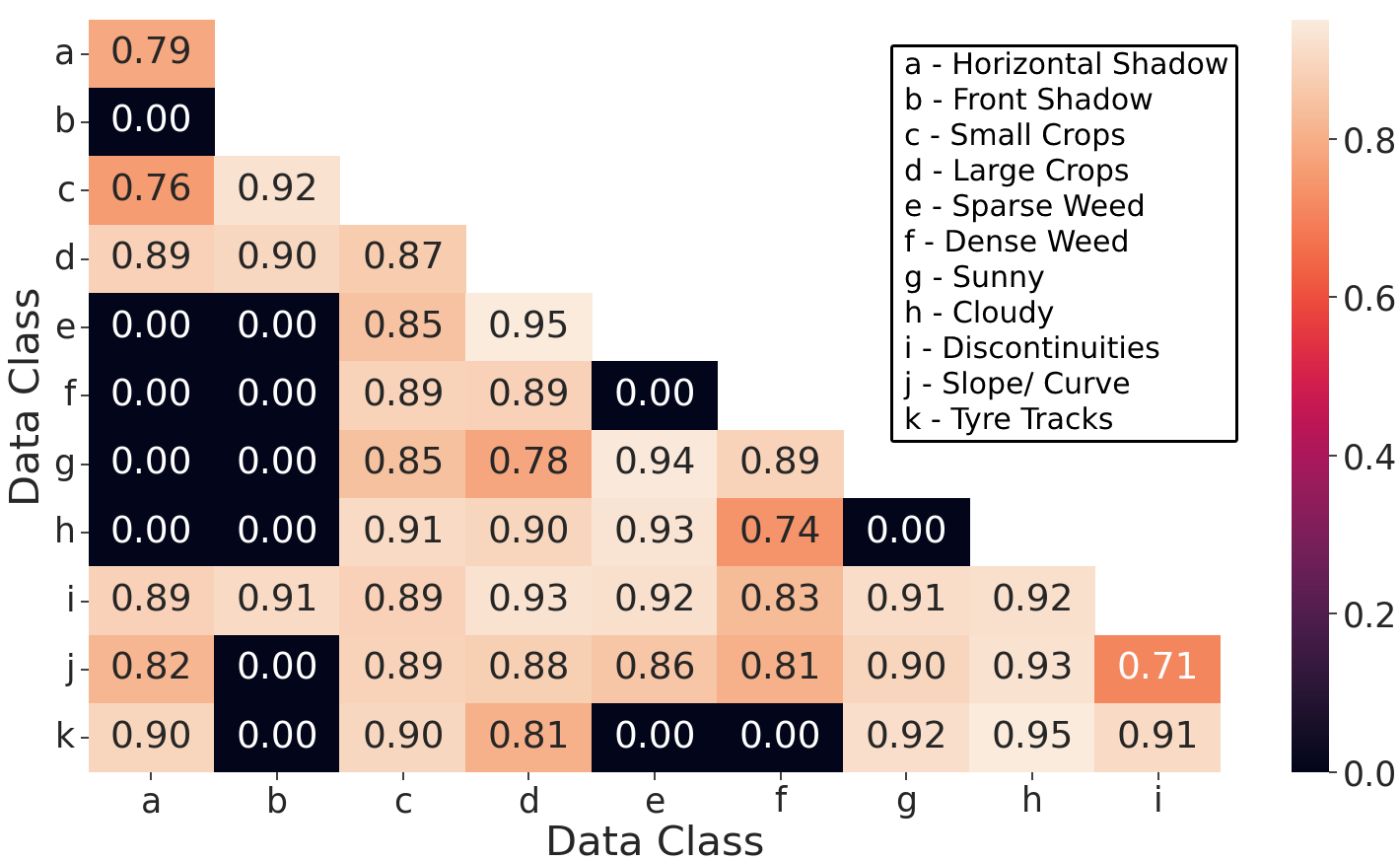}
\caption{Heatmap of Crop Row Detection Performance in Sugar Beet Field Variations (Zeros indicates the non-existing field variations)}
\label{fig:hms}
\end{figure}

\begin{figure}[t]
\centering
\captionsetup{justification=centering}
\includegraphics[scale=0.21]{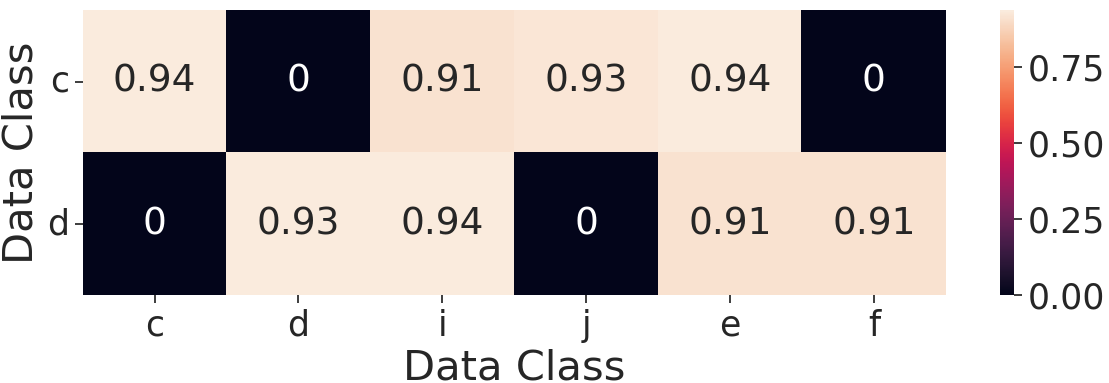}
\caption{Heatmap of Crop Row Detection Performance in Maize Field Variations (Zeros indicates the non-existing field variations)}
\label{fig:hmm}
\end{figure}

The evaluation of crop row detection is conducted based on the $\epsilon$ score described in Section \ref{sec:eps}. The heatmaps illustrated in Figure \ref{fig:hms} and Figure \ref{fig:hmm} depict the crop row detection performance of data classes belonging to Sugar Beet and Maize crops respectively. The average $\epsilon$ score for the entire dataset in our method was 90.25\%. The baseline method recorded an average $\epsilon$ score of 52.6\%. 

\begin{table}[t]
\centering
\caption{Crop Row Detection Performance ($\epsilon$) under Field Variations}
\begin{center}
\begin{tabular}{|p{0.3\linewidth} | p{0.12\linewidth} |}
\hline
\textbf{Category Name} & \textbf{$\epsilon$ Score} \\
\hline
{Horizontal Shadow (a)} & 83.88\%  \\ 
\hline
{Front Shadow (b)} & 91.32\%  \\ 
\hline
{Small Crops (c)} & 88.06\%  \\ 
\hline
{Large Crops (d)} & 90.41\%  \\ 
\hline
{Sparse Weed (e)} & 91.25\%  \\ 
\hline
{Dense Weed (f)} & 87.24\%  \\ 
\hline
{Sunny (g)} & 88.74\%  \\ 
\hline
{Cloudy (h)} & 90.1\%  \\ 
\hline
{Discontinuities (i)} & 87.79\%  \\ 
\hline
{Slope/ Curve (j)} & 88.67\%  \\ 
\hline
{Tyre Tracks (k)} & 89.69\%  \\ 
\hline
\end{tabular}
\label{tab:cep}
\end{center}
\end{table}

The data class which comprises curved crop rows with discontinuities (class 42) has the worst performance of crop row detection in our method with an $\epsilon$ score of 71.11\%. The best case crop row detection of the baseline was recorded in class 8 with an $\epsilon$ score of 72.22\%. The best crop row detection performance from our method was detected in class 41 which is comprised of crops near tramlines with cloudy overcast. The average crop row detection performance for Sugar Beet and Maize were 87.7\% and 92.8\% respectively. The baseline method failed to detect any crop rows in 10.4\% of the test cases whereas our method successfully detected crop rows in all the test images. The $\epsilon$ scores of data classes belonging to each of the 11 field variations were averaged to identify the impact of field variations on crop row detection. The average $\epsilon$ score for each field variation is listed in Table \ref{tab:cep}.

The least challenging field variations for our algorithm were the "Front Shadow" and "Sparse Weed" categories. The "Horizontal Shadow" and "Dense Weed" were the most challenging field variations for our method according to the results indicated in Table \ref{tab:cep}. However, a closer examination of the results in the "Horizontal Shadow" field variation revealed that an image in data class 2 had detected a false crop row causing such a lower score. The overall crop row detection in this category is similar to other categories if the borderline case was ignored. The false crop row detection had occurred in an image where the human driver accidentally drove the robot outside the crop row. This reveals a co-dependency between the accuracy of the crop row detection algorithm and the visual servoing controller. The crop row detection algorithm's accuracy is maintained as long as the visual servoing controller can maintain the robot within the crop row without causing major orientation swings. Our evaluations of the visual servoing controller prove that it can steer the robot accurately within the crop row. The addition of more images from the challenging field variations during the U-Net training could improve the performance of our algorithm against those conditions.

\subsection{Visual Servoing Evaluation}
\begin{figure}[t]
\centering
\captionsetup{justification=centering}
\includegraphics[scale=0.32]{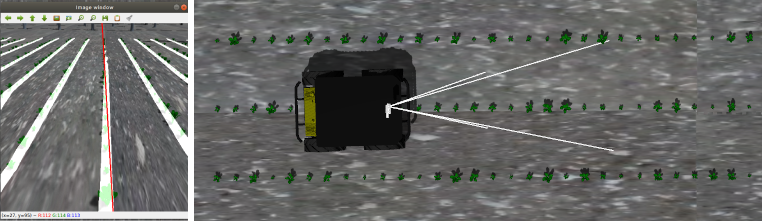}
\caption{Simulated Sugar Beet Field for Visual Servoing}
\label{fig:sim}
\end{figure}

\begin{figure}[t]
\centering
\captionsetup{justification=centering}
\includegraphics[scale=0.15]{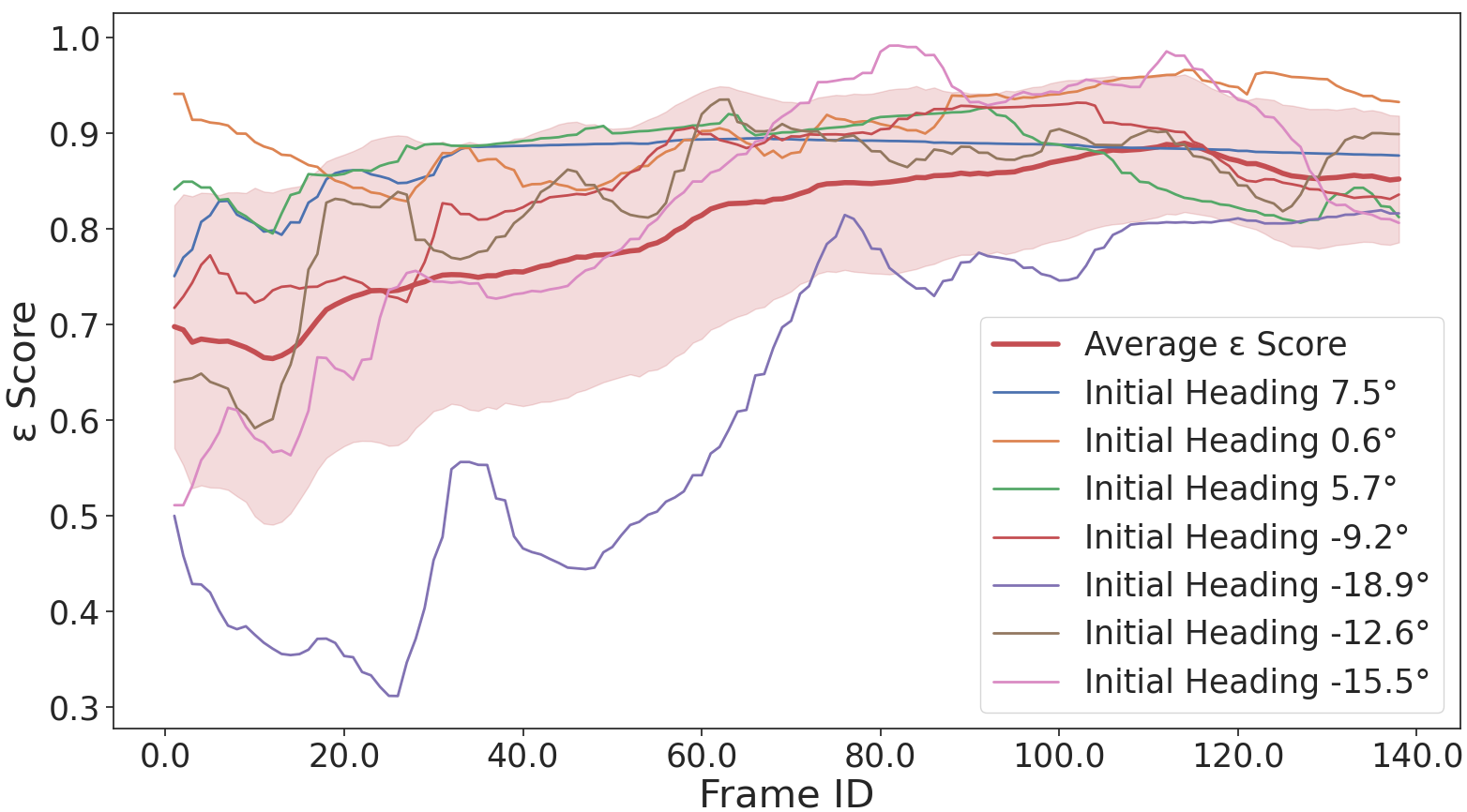}
\caption{$\epsilon$ Score Variation During Selected Trials of Visual Servoing (Red curve is the average score on all 20 trials)}
\label{fig:vsre}
\end{figure}

The performance of the visual servoing controller was evaluated in a simulation environment with a U-Net model trained on simulated data. The crop rows in the simulation were modelled in perfect 6m straight lines to observe the performance of our visual servoing controller. The robot was placed at the beginning of each crop row with a random initial heading angle within $\pm$20\textdegree. Figure \ref{fig:sim} shows the simulation environment and during a trial run of visual servoing. The $\epsilon$ score of the crop row detection algorithm was recorded to evaluate the performance of the visual servoing controller. Figure \ref{fig:vsre} shows the $\epsilon$ score curves during a few trials. The red colour curve in Figure \ref{fig:vsre} denotes the average $\epsilon$ score with standard deviation bands. Each trial took around 140 frames to execute. The average starting $\epsilon$ score was 69.76\% and it improved up to an average terminating $\epsilon$ score of 85.2\%. Large swings could be seen in the $\epsilon$ score curve corresponding to the initial angle of -18.9\textdegree due to the overshoots in the proportional controller. However, other $\epsilon$ score curves gradually settle to 80\% to 90\% range within 50 frames of execution.

\subsection{Execution of Exit Manoeuvre}
\begin{figure}[t]
\centering
\captionsetup{justification=centering}
\includegraphics[scale=0.15]{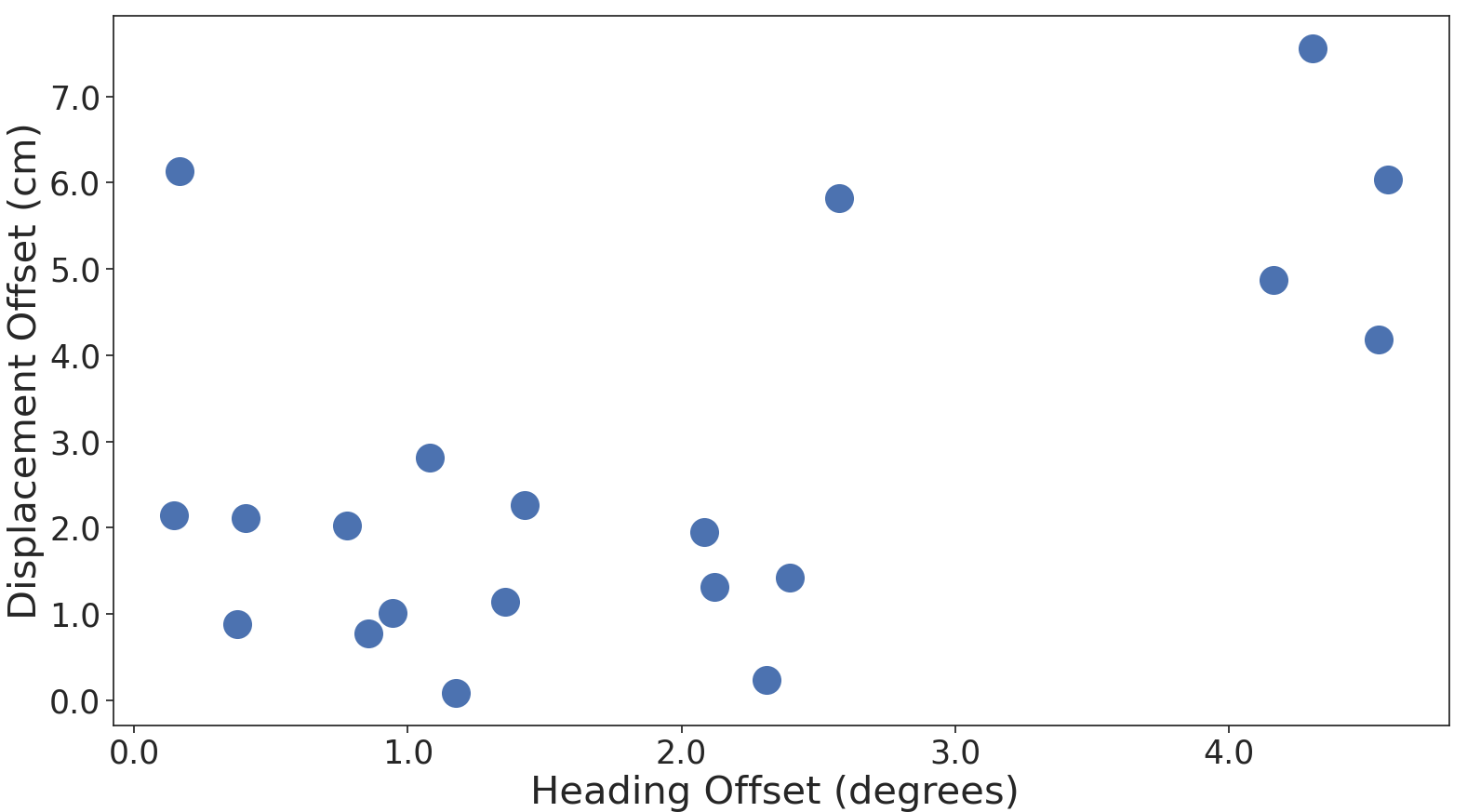}
\caption{Final Heading and Displacement Offsets After Executing the Exit Maneuver}
\label{fig:em}
\end{figure}

We set up an experiment to evaluate the ability of the exit manoeuvre to successfully guide the robot out of the crop row. The robot was allowed to follow the crop row and perform an exit manoeuvre for 20 trials in 20 simulated crop rows. The positional and angular offset of the robot was measured when the robot came to a halt after following the exit manoeuvre described in Section \ref{ssec:em}.  The $\lambda$ was set to 0.01 during the experiment. Figure \ref{fig:em} illustrates the distribution of the final heading and displacement offsets of the robot after exiting the crop row. The maximum recorded heading angle and displacement offsets were 4.55\textdegree and 7.56 cm. The average heading angle offset was 1.89\textdegree while the average displacement offset was 2.74 cm. These results indicate that the exit manoeuvre could be executed without damaging the crops to successfully exit the crop row. 

\section{Conclusion}
\label{sec:con}
We have presented a novel crop row detection algorithm for visual servoing along a crop row in an arable field. Our algorithm will detect crop rows with an average $\epsilon$ score of 90.25\% when the visual servoing controller maintains the $\Delta \theta$ with $\pm$20\textdegree. The overall crop row detection performance of our method is 37.66\% better than the baseline. We have identified that the dense population of weeds in inter-row spaces and discontinuities in crop rows are the most challenging field conditions for crop row detection. We also developed an EOR detector that detects the end of each crop row and guides the robot safely towards the headland area while exiting the crop row.  The EOR detector can navigate the robot to exit a crop row while maintaining a maximum final heading offset of 4.55\textdegree and staying less than 8cm distance away from the crop row position. Our method eliminates the need for additional hardware such as multiple cameras or variable FOV cameras for exiting the crop row. The work presented in this paper only covers the navigation of a robot along a single crop row. We will expand this work to enable the robot to switch crop rows and hence navigate the entire field. The visual servoing controller and the exit manoeuvre were only tested in a simulation environment. Although the results are promising under such controlled conditions, real-world testing must be done to validate its performance.  

\bibliography{root}

\begin{thebibliography}{10}
\providecommand{\url}[1]{#1}
\csname url@samestyle\endcsname
\providecommand{\newblock}{\relax}
\providecommand{\bibinfo}[2]{#2}
\providecommand{\BIBentrySTDinterwordspacing}{\spaceskip=0pt\relax}
\providecommand{\BIBentryALTinterwordstretchfactor}{4}
\providecommand{\BIBentryALTinterwordspacing}{\spaceskip=\fontdimen2\font plus
\BIBentryALTinterwordstretchfactor\fontdimen3\font minus
  \fontdimen4\font\relax}
\providecommand{\BIBforeignlanguage}[2]{{%
\expandafter\ifx\csname l@#1\endcsname\relax
\typeout{** WARNING: IEEEtran.bst: No hyphenation pattern has been}%
\typeout{** loaded for the language `#1'. Using the pattern for}%
\typeout{** the default language instead.}%
\else
\language=\csname l@#1\endcsname
\fi
#2}}
\providecommand{\BIBdecl}{\relax}
\BIBdecl

\bibitem{oliveira2021advances}
L.~F. Oliveira, A.~P. Moreira, and M.~F. Silva, ``Advances in agriculture
  robotics: A state-of-the-art review and challenges ahead,'' \emph{Robotics},
  vol.~10, no.~2, p.~52, 2021.

\bibitem{bonadies2019overview}
S.~Bonadies and S.~A. Gadsden, ``An overview of autonomous crop row navigation
  strategies for unmanned ground vehicles,'' \emph{Engineering in Agriculture,
  Environment and Food}, vol.~12, no.~1, pp. 24--31, 2019.

\bibitem{adhikari2020deep}
S.~P. Adhikari, G.~Kim, and H.~Kim, ``Deep neural network-based system for
  autonomous navigation in paddy field,'' \emph{IEEE Access}, vol.~8, pp.
  71\,272--71\,278, 2020.

\bibitem{pang2020improved}
Y.~Pang, Y.~Shi, S.~Gao, F.~Jiang, A.-N. Veeranampalayam-Sivakumar,
  L.~Thompson, J.~Luck, and C.~Liu, ``Improved crop row detection with deep
  neural network for early-season maize stand count in uav imagery,''
  \emph{Computers and Electronics in Agriculture}, vol. 178, p. 105766, 2020.

\bibitem{bah2019crownet}
M.~D. Bah, A.~Hafiane, and R.~Canals, ``Crownet: Deep network for crop row
  detection in uav images,'' \emph{IEEE Access}, vol.~8, pp. 5189--5200, 2019.

\bibitem{ji2011crop}
R.~Ji and L.~Qi, ``Crop-row detection algorithm based on random hough
  transformation,'' \emph{Mathematical and Computer Modelling}, vol.~54, no.
  3-4, pp. 1016--1020, 2011.

\bibitem{fue2020evaluation}
K.~Fue, W.~Porter, E.~Barnes, C.~Li, and G.~Rains, ``Evaluation of a stereo
  vision system for cotton row detection and boll location estimation in direct
  sunlight,'' \emph{Agronomy}, vol.~10, no.~8, p. 1137, 2020.

\bibitem{ahmadi2020visual}
A.~Ahmadi, L.~Nardi, N.~Chebrolu, and C.~Stachniss, ``Visual servoing-based
  navigation for monitoring row-crop fields,'' in \emph{2020 IEEE International
  Conference on Robotics and Automation (ICRA)}.\hskip 1em plus 0.5em minus
  0.4em\relax IEEE, 2020, pp. 4920--4926.

\bibitem{romeo2012crop}
J.~Romeo, G.~Pajares, M.~Montalvo, J.~Guerrero, M.~Guijarro, and A.~Ribeiro,
  ``Crop row detection in maize fields inspired on the human visual
  perception,'' \emph{The Scientific World Journal}, vol. 2012, 2012.

\bibitem{guerrero2013automatic}
J.~M. Guerrero, M.~Guijarro, M.~Montalvo, J.~Romeo, L.~Emmi, A.~Ribeiro, and
  G.~Pajares, ``Automatic expert system based on images for accuracy crop row
  detection in maize fields,'' \emph{Expert Systems with Applications},
  vol.~40, no.~2, pp. 656--664, 2013.

\bibitem{woebbecke1995color}
D.~M. Woebbecke, G.~E. Meyer, K.~Von~Bargen, and D.~A. Mortensen, ``Color
  indices for weed identification under various soil, residue, and lighting
  conditions,'' \emph{Transactions of the ASAE}, vol.~38, no.~1, pp. 259--269,
  1995.

\bibitem{bakker2008vision}
T.~Bakker, H.~Wouters, K.~Van~Asselt, J.~Bontsema, L.~Tang, J.~M{\"u}ller, and
  G.~van Straten, ``A vision based row detection system for sugar beet,''
  \emph{Computers and electronics in agriculture}, vol.~60, no.~1, pp. 87--95,
  2008.

\bibitem{montalvo2012automatic}
M.~Montalvo, G.~Pajares, J.~M. Guerrero, J.~Romeo, M.~Guijarro, A.~Ribeiro,
  J.~J. Ruz, and J.~Cruz, ``Automatic detection of crop rows in maize fields
  with high weeds pressure,'' \emph{Expert Systems with Applications}, vol.~39,
  no.~15, pp. 11\,889--11\,897, 2012.

\bibitem{emmi2022toward}
L.~Emmi., J.~Herrera{-}Diaz., and P.~Gonzalez{-}de{-}Santos., ``Toward
  autonomous mobile robot navigation in early-stage crop growth,'' in
  \emph{Proceedings of the 19th International Conference on Informatics in
  Control, Automation and Robotics - ICINCO,}, INSTICC.\hskip 1em plus 0.5em
  minus 0.4em\relax SciTePress, 2022, pp. 411--418.

\bibitem{fawakherji2019crop}
M.~Fawakherji, A.~Youssef, D.~Bloisi, A.~Pretto, and D.~Nardi, ``Crop and weeds
  classification for precision agriculture using context-independent pixel-wise
  segmentation,'' in \emph{2019 Third IEEE International Conference on Robotic
  Computing (IRC)}.\hskip 1em plus 0.5em minus 0.4em\relax IEEE, 2019, pp.
  146--152.

\bibitem{de2022towards}
R.~de~Silva, G.~Cielniak, and J.~Gao, ``Towards infield navigation: leveraging
  simulated data for crop row detection,'' \emph{arXiv preprint
  arXiv:2204.01811}, 2022.

\bibitem{winterhalter2018crop}
W.~Winterhalter, F.~V. Fleckenstein, C.~Dornhege, and W.~Burgard, ``Crop row
  detection on tiny plants with the pattern hough transform,'' \emph{IEEE
  Robotics and Automation Letters}, vol.~3, no.~4, pp. 3394--3401, 2018.

\bibitem{GAO201843}
J.~Gao, W.~Liao, D.~Nuyttens, P.~Lootens, J.~Vangeyte, A.~Pi{\v{z}}urica,
  Y.~He, and J.~G. Pieters, ``Fusion of pixel and object-based features for
  weed mapping using unmanned aerial vehicle imagery,'' \emph{International
  journal of applied earth observation and geoinformation}, vol.~67, pp.
  43--53, 2018.

\bibitem{winterhalter2021localization}
W.~Winterhalter, F.~Fleckenstein, C.~Dornhege, and W.~Burgard, ``Localization
  for precision navigation in agricultural fields—beyond crop row
  following,'' \emph{Journal of Field Robotics}, vol.~38, no.~3, pp. 429--451,
  2021.

\bibitem{xue2012variable}
J.~Xue, L.~Zhang, and T.~E. Grift, ``Variable field-of-view machine vision
  based row guidance of an agricultural robot,'' \emph{Computers and
  Electronics in Agriculture}, vol.~84, pp. 85--91, 2012.

\bibitem{ahmadi2021towards}
A.~Ahmadi, M.~Halstead, and C.~McCool, ``Towards autonomous crop-agnostic
  visual navigation in arable fields,'' \emph{arXiv preprint arXiv:2109.11936},
  2021.

\bibitem{desilva22deep}
\BIBentryALTinterwordspacing
R.~de~Silva, G.~Cielniak, G.~Wang, and J.~Gao, ``Deep learning-based crop row
  following for infield navigation of agri-robots,'' 2022. [Online]. Available:
  \url{https://arxiv.org/abs/2209.04278}
\BIBentrySTDinterwordspacing

\bibitem{ronneberger2015u}
O.~Ronneberger, P.~Fischer, and T.~Brox, ``U-net: Convolutional networks for
  biomedical image segmentation,'' in \emph{International Conference on Medical
  image computing and computer-assisted intervention}.\hskip 1em plus 0.5em
  minus 0.4em\relax Springer, 2015, pp. 234--241.

\bibitem{espiau1992new}
B.~Espiau, F.~Chaumette, and P.~Rives, ``A new approach to visual servoing in
  robotics,'' \emph{ieee Transactions on Robotics and Automation}, vol.~8,
  no.~3, pp. 313--326, 1992.

\end{thebibliography}
\end{document}